\documentclass{article}

\usepackage{PRIMEarxiv}
\usepackage{amsmath}
\usepackage[noend]{algorithmic}
\usepackage{algorithm}
\usepackage{subfigure}
\usepackage{subcaption}

\usepackage[utf8]{inputenc} 
\usepackage[T1]{fontenc}    
\usepackage{hyperref}       
\usepackage{url}            
\usepackage{booktabs}       
\usepackage{amsfonts}       
\usepackage{nicefrac}       
\usepackage{microtype}      
\usepackage{lipsum}
\usepackage{fancyhdr}       
\usepackage{graphicx}       
\graphicspath{{media/}}     

\title{Dyna-Style Learning with A Macroscopic Model for Vehicle Platooning in Mixed-Autonomy Traffic
}

\author{
  Yichuan Zou \\
  Shanghai Jiao Tong University \\
  \texttt{zouyichuan@sjtu.edu.cn} \\
  \And
  Li Jin \\
  Shanghai Jiao Tong University \\
  \texttt{li.jin@sjtu.edu.cn} \\
  \And
  Xi Xiong \\
  Tongji University \\
  \texttt{xi\_xiong@tongji.edu.cn} \\
}

\begin{document}
\maketitle

\begin{abstract}
Platooning of connected and autonomous vehicles (CAVs) plays a vital role in modernizing highways, ushering in enhanced efficiency and safety. This paper explores the significance of platooning in smart highways, employing a coupled partial differential equation (PDE) and ordinary differential equation (ODE) model to elucidate the complex interaction between bulk traffic flow and CAV platoons. Our study focuses on developing a Dyna-style planning and learning framework tailored for platoon control, with a specific goal of reducing fuel consumption. By harnessing the coupled PDE-ODE model, we improve data efficiency in Dyna-style learning through virtual experiences. Simulation results validate the effectiveness of our macroscopic model in modeling platoons within mixed-autonomy settings, demonstrating a notable $10.11\%$ reduction in vehicular fuel consumption compared to conventional approaches.
\end{abstract}


\section{Introduction}

Traffic congestion presents a formidable obstacle, arising when traffic demand surpasses road capacity, leading to prolonged wait times and heightened fuel consumption due to frequent acceleration and deceleration maneuvers~\cite{arnott1994economics}. Highways, with their inherent complexities like on-ramps and lane reductions, often experience exacerbated congestion scenarios. In response, platooning—coordinated movement of connected and autonomous vehicles (CAVs)—emerges as a promising strategy for optimizing traffic flow and mitigating congestion.

\begin{figure}[htbp]
\centering
\includegraphics[width=0.5\textwidth, trim=110 230 88 170,clip]{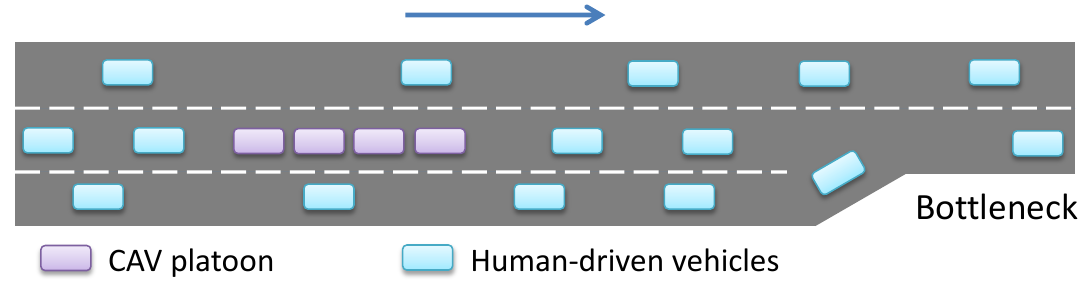}
\caption{The mixed-autonomy traffic through a bottleneck on highway segment.}
\label{Platoon at bottleneck}
\end{figure}

This study focuses on leveraging vehicle platooning to alleviate traffic congestion at bottlenecks in mixed autonomy settings, as illustrated in Fig.~\ref{Platoon at bottleneck}. Platooning management encompasses two main paradigms: model-based and model-free approaches. Model Predictive Control (MPC) stands out as a prevalent model-based method, offering flexibility and adaptability in platoon management. Studies by Wang et al.~\cite{wang2021model}, Feng et al.~\cite{feng2021robust}, and Hu et al.~\cite{hu2022fuel} underscore MPC's efficacy in optimizing platoon behavior while ensuring fuel efficiency and minimal traffic disruption across varying environmental conditions. Another model-based approach involves the use of optimal control techniques like Linear Quadratic Regulator (LQR) and Model Predictive Path Integral Control (MPPI), which have shown promising results in platoon management. Research by Stankovic et al.~\cite{stankovic2000decentralized} and Uhlemann~\cite{uhlemann2016platooning} demonstrates the effectiveness of these methods in optimizing platoon behavior and reducing fuel consumption. Additionally, reinforcement learning (RL) offers a model-free approach, where agents interact with their environment to learn optimal action policies. Notably, Li et al.~\cite{li2021reinforcement} and Liu et al.~\cite{liu2022autonomous} showcase RL's effectiveness in platoon control, emphasizing its ability to learn policies that maximize cumulative rewards, thereby optimizing traffic flow and reducing fuel consumption in complex traffic scenarios.

In this work, we present a novel contribution to the field by introducing a Dyna-style learning framework rooted in a first-order traffic flow model, aimed at effectively managing platoons and reducing vehicular fuel consumption during lane reductions. At the core of our approach lies the traditional macroscopic Lighthill-Whitham-Richards (LWR) model~\cite{lighthill1955kinematic, richards1956shock}, which consists of partial differential equations (PDEs) capturing traffic flow dynamics and ordinary differential equations (ODEs) tracking platoon trajectories. Through parameterization and Kalman filter updates~\cite{kalman1960new, xiong2020dynamic}, our traffic model dynamically adapts to the environment, enabling accurate representation of forward dynamics. Unlike conventional neural network-based models prevalent in model-based RL, our parameterized traffic model offers superior interpretability and requires fewer parameters, owing to its foundation in physical fluid flow principles.

Furthermore, we evaluate our proposed framework within the Simulation of Urban MObility (SUMO) platform~\cite{behrisch2011sumo}. Results demonstrate that our framework enhances data efficiency and accelerates training compared to model-free learning algorithms. Under our framework's policy control, fuel consumption significantly decreases compared to the Krauss model~\cite{krauss1998microscopic}, the default car-following model in SUMO.

The rest of this paper is organized as follows.
Section~\ref{sec:model} delves into the PDE-ODE traffic model, detailing its update and prediction mechanisms.
Section~\ref{sec:RL} introduces our Dyna-style planning and learning framework. 
In Section~\ref{sec:simulation}, we meticulously evaluate our model's accuracy and policy performance. 
Finally, Section~\ref{sec:conclusion} summarizes our key findings and outlines future research directions.

\section{Forward dynamics with a macroscopic model\label{sec:model}}

\subsection{The coupled PDE-ODE model}\label{sec:model:intro}

In this section, we model traffic flow on a multi-lane highway using a hyperbolic partial differential equation (PDE) based on the LWR model. Trajectories of platoons are tracked with an ordinary differential equation (ODE), assuming a fixed platoon length. Here, $\rho(x, t)$ represents vehicle density, $F(t,x,\rho)$ is the flow function, and $v=v(\rho)$ is the average speed.
The platoon's downstream endpoint is denoted by $z_d(t)$, with initial conditions $\rho_0(x)$ and $z_d^0$ \cite{piacentini2020macroscopic}. 
The model is summarized as:
\begin{subequations} 
\label{}
\begin{align}
\label{1a}
&\partial_{t} \rho + \partial_{x} F(t,x,\rho) = 0,\\
&\rho(0,x) = \rho_0(x),\\
&\dot{z}_d(t) = v_d (t, \rho(t, z_d(t)+)),\\
&z_d(0) = z_d^0.
\end{align}
\end{subequations}
We adopt a linear density-speed relation expressed as:
\begin{equation}\label{density-speed}
v(\rho)=V\left(1-\frac{\rho}{R}\right),
\end{equation}
where $V$ represents the maximum speed of vehicles, and $R$ denotes the maximal reachable density on the road. The flow function $F(t,x,\rho)$ is defined as $f(\rho) = V \rho \left(1 - \frac{\rho}{R} \right)$.
To determine the speed of the downstream point, we establish:
\begin{equation*}
    v_d(t, \rho) = \min \{V_d(t), v(\rho)\},
\end{equation*}
where $V_d(t) \in [0, V ] $ represents the maximum speed of $z_d(t)$. The platoon is considered as a rigid body which has a fixed length. The speed of overall platoon is defined as $v_{p}$, and we have $v_{p} = v_{d}$.

To solve the PDE in Eq.~(\ref{1a}), we discretize both time and space, sampling with a time interval $\Delta t$ and dividing the highway into cells with length $\Delta x$. The time interval and cell length should satisfy the Courant-Friedrichs-Lewy (CFL) condition, ensuring stability: $2V \Delta t \le \Delta x$ \cite{courant1967partial}.
The update of density in each cell $i$ at time step $k$ is given by:
\begin{equation} 
\label{densityChange}
    \rho_{i}(k+1) = \rho_{i}(k)-\frac{\Delta t}{\Delta x}\left(F_{i+1/2}(k)-F_{i-1/2}(k)\right),
\end{equation}
where $F_{i+1/2}$ represents the flow between cell $i$ and cell $i+1$, computed using demand and supply functions.
The demand and supply of a cell are computed as follows:
\begin{subequations}
    \begin{align*}
        D_{i}(k) =\left\{\begin{array}{ll}
        f\left(\rho_{i}(k)\right) & \text { if } \rho_{i}(k)<\rho_{c r}, \\
        f_{\max} & \text { if } \rho_{i}(k) \geq \rho_{c r},
        \end{array}\right. \quad
        S_{i}(k) =\left\{\begin{array}{ll}
        f_{\max} & \text { if } \rho_{i}(k)<\rho_{c r}, \\
        f\left(\rho_{i}(k)\right) & \text { if } \rho_{i}(k) \geq \rho_{c r},
        \end{array}\right.
    \end{align*}
\end{subequations}
where $\rho_{c r} = 0.5R$ denotes the critical density and $f_{\max}$ is the maximal flow between cells. 
The flow between cells is determined as the minimum between the demand of the upstream cell and the supply of the downstream cell:
\begin{equation*}
    F_{i}(k)=\min \left\{D_{i-1}(k), S_{i}(k)\right\}.
\end{equation*}

\subsection{One-step prediction with Kalman filter}

Kalman filtering is a powerful technique for estimating the true state of a system in the presence of noisy measurements \cite{kalman1960new}. The Kalman filter framework is structured as follows:

\begin{subequations} 
\label{Kalman}
    \begin{align} 
    \label{7a}
    \hat{\phi}_{k}^{\ominus}&=F_{k-1} \hat{\phi}_{k-1}^{\oplus}+G_{k-1} u_{k-1}, \\
    P_{k}^{\ominus}&=F_{k-1} P_{k-1}^{\oplus} F_{k-1}^{T}+Q_{k-1}, \\
    K_{k}&=P_{k}^{\ominus} h_{k}\left(r_{k}+h_{k}^{\top} P_{k}^{\ominus} h_{k}\right)^{-1}, \\
    P_{k}^{\oplus}&=\left(I-K_{k} h_{k}^{\top}\right) P_{k}^{\ominus}, \\
    \hat{\phi}_{k}^{\oplus}&=\hat{\phi}_{k}^{\ominus}+K_{k}\left(y_{k}-h_{k}^{\top} \hat{\phi}_{k}^{\ominus}\right),
    \end{align} 
\end{subequations}
where $\hat{\phi}_{k}^{\ominus}$ represents the predicted state at step $k$, incorporating the dynamics of the system $F_k$ and any control inputs $G_{k-1} u_{k-1}$ from the previous updated state $\hat{\phi}_{k-1}^{\oplus}$.
The covariance matrix $P_{k}^{\ominus}$ captures the uncertainty associated with the predicted state, accounting for both the uncertainty carried over from the previous step $P_{k-1}^{\oplus}$ and the process noise $Q_{k-1}$.
The Kalman gain $K_{k}$ adjusts the predicted state based on the discrepancy between the observed state $y_{k}$ and the predicted state, weighted by the measurement noise $r_{k}$ and the transformation matrix $h_{k}$.
Updating the covariance matrix via $P_{k}^{\oplus}$ refines the uncertainty estimate, adjusting it based on the Kalman gain and the measurement transformation.
Finally, the updated state $\hat{\phi}_{k}^{\oplus}$ is computed by adjusting the predicted state with the Kalman gain, aligning it with the observed measurements.

In our project, we aim to develop a comprehensive predictive model for the simulation environment, enabling the forecasting of both traffic density and average speed in the subsequent time step based on current information. This model consists of two primary components: firstly, predicting density in the next time step by utilizing a traffic flow model, where we calculate the flow rate of each cell using demand and supply functions, and then applying a recursion formula derived from the PDE represented by Eq. (\ref{densityChange}) to update the density iteratively. Secondly, we focus on predicting speed in the next time step by leveraging the projected density from the previous step and coupling it with appropriate speed models.

For the second part, building on the PDE-ODE model's insight into the linear relationship between density and speed, we approach the task by treating $V$ (maximal speed) and $R$ (maximal density) as parameters that require adaptation to fit the environment. To achieve this, we employ a Kalman filter to iteratively update these parameters. In this adaptation, the state $\phi$ in the Kalman filter represents a two-dimensional vector encapsulating $V$ and $R$. Given that maximal speed and maximal density are inherent attributes of road sections, the status transition matrix in Eq.~(\ref{7a}) simplifies to the identity matrix ($F_k=I$), and the control matrix becomes a zero matrix ($G_k=0$), with no process noise ($Q_k=0$) introduced. For simplicity, we define the weight $r_k$ as 1. Consequently, the Kalman filter reduces to the following form:
\begin{subequations} \label{Simplified_Kalman}
    \begin{align} 
    K_{k}&=P_{k-1} h_{k}\left(1+h_{k}^{\top} P_{k-1} h_{k}\right)^{-1}, \\
    P_{k}&=\left(I-K_{k} h_{k}^{\top}\right) P_{k-1}, \\
    \label{9c}
    {\phi}_{k}&={\phi}_{k-1}+K_{k}\left(y_{k}-h_{k}^{\top} {\phi}_{k-1}\right).
    \end{align} 
\end{subequations}

Consider a specific road cell where we observe density $\rho_k$ and average speed $v_k$ at step $k$ in the simulation process.
Notably, the observed speed coincides with the actual cell speed. In Eq.~(\ref{9c}), we define $y_k = v_k$. $R_{k-1}$ and $V_{k-1}$ denote road attributes obtained from the previous step. By defining the linear transformation $h_{k}=[1, -\rho_k]$ and initializing the status $\phi_{k-1}=[V_{k-1}, \frac{V_{k-1}}{R_{k-1}}]$, we can predict the average speed of the cell by the inner product of these two vectors:
\begin{equation*}
    \overline{v}_{k} = h_{k}^{\top} {\phi}_{k-1} = V_{k-1}(1-\frac{\rho_{k}}{R_{k-1}}).
\end{equation*}
Following the update in Eq.~(\ref{9c}), we acquire an updated parameter set $\phi_{k}=[V_{k}, \frac{V_{k}}{R_{k}}]$. By iterating through the steps of the simplified Kalman filter, these model parameters undergo continuous refinement and updating, ensuring the model's adaptability and accuracy in predicting the dynamics of the simulation environment.

To enhance the accuracy and flexibility of our model, we can adapt it to accommodate varying road conditions, particularly in regions with bottlenecks.
This involves dividing the entire road into several sections of equal length and employing separate parameters for each section, thereby tailoring the model to the specific characteristics of each segment.
With the status of CAV platoons denoted as $[x_{p}(k), v_{p}(k)]$, alongside cell density $\rho_{i}(k)$ and cell speed $\overline{v}_i(k)$ at step $k$, the prediction process for step $k+1$ unfolds as follows:
\begin{enumerate}
    \item Platoon status prediction: 
    \begin{align*}
        v_{p}(k+1) &= v_{p}(k) + a_{p}(k) * \Delta t,\\
        x_{p}(k+1) &= x_{p}(k) + v_{p}(k) * \Delta t + \frac{1}{2}a_{p}(k) * \Delta t.
    \end{align*}
    
    \item Cell density prediction: 
    \begin{align*}
        & F_{i}(k) = \min \left\{D_{i-1}(k), S_{i}(k)\right\},\\
        & \rho_{i}(k+1) = \rho_{i}(k)-\frac{\Delta t}{\Delta x}\left(F_{i+1 / 2}(k)-F_{i-1 / 2}(k)\right),
    \end{align*}
    
    \item Cell speed prediction:
    \begin{align*}
        \overline{v_{i}}(k+1)=V_{n}(1-\frac{\rho_{i}(k+1)}{R_{n}}),
    \end{align*}
    
\end{enumerate}
where $i$ represents the cell index and $n$ denotes the segment index. This segmentation approach enables our model to adapt more effectively to diverse road conditions.

\section{Dyna-Style Planning and Learning}\label{sec:RL}

In this section, we introduce a Markov Decision Process (MDP) framework for managing vehicle platoons within a mixed autonomy setting. Leveraging a Dyna-style approach, our framework seamlessly integrates model-free learning with model-based planning to govern platoon behavior. The model-based planning component relies on insights gleaned from the coupled PDE-ODE model outlined in the preceding section, allowing for a comprehensive understanding of platoon dynamics and informed decision-making.

\subsection{Markov decision process}

The CAV platoon aims to learn an optimal action policy that minimizes fuel consumption within the simulation environment. To formalize this objective, we define a Markov Decision Process (MDP) model denoted as $M = {\mathcal{S}, \mathcal{A}, \mathcal{P}, \mathcal{R}, \gamma}$. Here, $\mathcal{S}$ represents the set of states, $\mathcal{A}$ denotes the set of actions, and $\mathcal{P}: \mathcal{S} \times \mathcal{A} \to [0, 1]$ characterizes the dynamics determined by the simulation environment. At a given state $s \in \mathcal{S}$, the platoon selects an action $a \in \mathcal{A}$ based on its policy and transitions to a new state $s^\prime \in \mathcal{S}$ while receiving a reward $r \in \mathcal{R}$.

$\textbf{State} \quad$ The state is represented by a vector capturing various aspects of the environment. Assuming the entire road section is divided into $N$ cells, each with a length of $L$ meters, the state vector includes dimensions for platoon position and speed, followed by densities and average speeds of all cells. This yields a state vector with $2+2N$ dimensions:
\begin{equation*}
    s_t = [x_{p}, v_{p}, \rho_{1}, ..., \rho_N, \bar{v}_{1}, ..., \bar{v}_N].
\end{equation*}

$\textbf{Action} \quad$ The platoon adjusts its acceleration as its action. The action space is discrete: $a_t \in \{a|a_{\min} \le a \le a_{\max}, \ a \in \mathbb{N} \}$. If an action results in the platoon's speed becoming negative or exceeding the maximum speed, the platoon's speed is constrained to remain within the allowed range.

$\textbf{Reward} \quad$ The platoon's primary objective is to traverse the entire road section while minimizing total vehicle fuel consumption. The average fuel consumption rate (L/s) is computed using a sixth-order polynomial based on the average vehicle speed, as described in ~\cite{ramadan2017traffic}. The polynomial equation is:\\
$$
\begin{array}{l}
    k_t(v)=5.7 \cdot 10^{-12} \cdot v^{6}-3.6 \cdot 10^{-9} \cdot v^{5}+7.6 \cdot 10^{-7} \cdot v^{4}-6.1 \cdot 10^{-5} \cdot v^{3}+1.9 \cdot 10^{-3} \cdot v^{2}\\+1.6 \cdot 10^{-2} \cdot v+0.99,
\end{array}
$$
From this equation, the total fuel consumption can be derived. In each time step, the platoon incurs a negative total fuel consumption penalty $r_{fc}$. Additionally, the platoon receives a fixed reward $r_{bonus}$ each time it enters the next road cell or reaches the end of the entire road section.
To ensure the platoon completes the entire road section within a limited time steps, it incurs a timeout penalty $r_{ot}$ based on the platoon's position.
An additional penalty $r_{acc}$ is applied to the platoon based on its acceleration. Consequently, the total reward is computed as $r_{t} = r_{fc} + r_{bonus} + r_{ot} + r_{acc}$.

\subsection{Dyna-Q learning with the macroscopic model}

Dyna-Q stands as a classical model-based reinforcement learning algorithm, utilizing Q-planning to generate experiences from a model and leveraging these experiences to update the policy \cite{sutton2018reinforcement}. While Dyna-Q represents a foundational approach in model-based reinforcement learning, our algorithm extends this framework by incorporating a novel parameterized PDE-ODE traffic model, a departure from the typical neural network-based models prevalent in contemporary research. This approach offers enhanced interpretability and requires fewer parameters compared to neural networks.

In our Dyna framework, actions are selected using an $\epsilon$-greedy policy with the Q network $Q_{\theta}$.
At each step, the agent transitions to the next state based on the selected action, with state transitions stored in replay memory $\mathcal{D}$.
Simultaneously, the Q network $Q_{\theta}$ is updated with the TD error, while the parameters of the traffic model $W_{\phi}$ are updated through a Kalman filter in in Eq.~(\ref{Simplified_Kalman}).
These parameters $\phi=[R_1, ..., R_n, V_1, ..., V_n]$ encapsulate the maximum density and cell speed of all segments.
Following the model update, $W_{\phi}$ predicts the next state and reward using sampled state-action pairs, and $Q_{\theta}$ is updated accordingly using these virtual experiences. The framework of our algorithm is presented in Algorithm~\ref{alg:Dyna-Q}.

\begin{algorithm}[htbp]
    \caption{Dyna-Q with the PDE-ODE model}
    \label{alg:Dyna-Q}
    \renewcommand{\algorithmicrequire}{\textbf{Input:}}
    \renewcommand{\algorithmicensure}{\textbf{Output:}}
    \begin{algorithmic}[1]
        \STATE  Initialize networks and model: $Q_{\theta}$, $Q_{\theta^{\prime}}$, $W_{\phi}$
        \STATE  Initialize replay memory: $\mathcal{D}$
        \FOR{episode = 1, M}
            \STATE  Initialize state $\textbf{s}_0$.
            \FOR{$t = 1, T$}
                \STATE Choose action $\textbf{a}_t$ using $\epsilon$-greedy($\textbf{s}_t$, $Q_{\theta}$)
                \STATE $\textbf{r}_t, \ \textbf{s}_{t+1}\leftarrow$ take action $\textbf{a}_t$ at state $\textbf{s}_t$
                \STATE $\mathcal{D} \leftarrow \mathcal{D} \cup \{(\textbf{s}_t, \textbf{a}_t, \textbf{r}_t, \textbf{s}_{t+1})\}$
                \STATE Sample state transitions: $(\textbf{s}, \textbf{a}, \textbf{r}, \textbf{s}^{\prime}) \sim \mathcal{D}$
                \STATE Update $Q_{\theta}$ with TD error:                 $L(\theta)=\left(Q_{\theta}(\textbf{s}, \textbf{a}) - \left(\textbf{r}+\max_{a^{\prime}}Q_{\theta^{\prime}}\left(\textbf{s}^{\prime}, \textbf{a}^{\prime} \right) \right) \right)^2$
                \STATE Update $W_{\phi}$ using the Kalman filter in Eq.~(\ref{Simplified_Kalman})
                \FOR{$c = 1, N$}
                    \STATE Sample state-action pairs: $(\textbf{s}, \textbf{a}) \sim \mathcal{D}$
                    \STATE $\textbf{r}_m, \ \textbf{s}^{\prime}_m \leftarrow W_{\phi}(\textbf{s}, \textbf{a})$
                    \STATE Update $Q_{\theta}$ with TD error:                     $L(\theta)=\left(Q_{\theta}(\textbf{s}, \textbf{a}) - \left(\textbf{r}_m+\max_{a^{\prime}}Q_{\theta^{\prime}}(\textbf{s}^{\prime}_m, \textbf{a}^{\prime}) \right) \right)^2$
                \ENDFOR
                \STATE Update target network every $C$ steps: $\theta^{\prime} \leftarrow \theta$
            \ENDFOR
        \ENDFOR
    \end{algorithmic}
\end{algorithm}

\section{Numerical results}\label{sec:simulation}

In this section, we provide an overview of our simulation environment setup. Subsequently, we evaluate the performance of the macroscopic model and Dyna-style policy within this environment. We delve into each aspect in the subsequent sections, systematically assessing the effectiveness and efficiency of our approach.

\subsection{Simulation settings}

\begin{figure*}[htbp]
\centering
\includegraphics[width=1\textwidth]{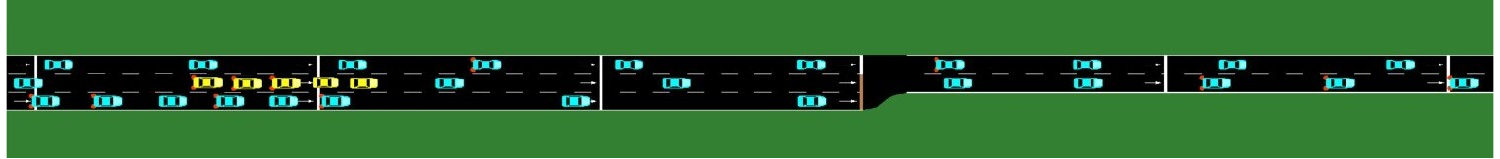}
\caption{The road section is divided into three sections: a preheating section, a three-lane section, and a two-lane section. The bottleneck occurs at the junction between the three-lane and two-lane sections.}
\label{bottleneck}
\end{figure*}

In the SUMO simulation environment, we construct a road section spanning a total length of 1000 meters, comprising three distinct segments: a 200-meter preheating section, a 500-meter three-lane section, and a 300-meter two-lane section. As depicted in Fig.~\ref{bottleneck}, the bottleneck emerges at the junction between the three-lane and two-lane sections, where congestion is anticipated to occur.

In this environment, we distinguish between two types of vehicles: CAV platoon and human-driven vehicles. The platoon is considered a distinct entity with a fixed length, longer than standard vehicles. Specifically, the lengths of the platoon and human-driven vehicles are set to $30$ meters and $5$ meters, respectively. Both platoon and human-driven vehicles operate with a maximum speed of $15$ meters per second, while the platoon's acceleration ranges from $-5$ to $3$ meters per second squared. The flow rate of human-driven vehicles is maintained at $3600$ per hour. The platoon enters the preheating section at the $100$ seconds, operating at maximum speed. Upon entering the three-lane section, our approach assumes control of the platoon's operations.

\subsection{Dyna-Q settings}

Based on the Courant-Friedrichs-Lewy (CFL) condition $2V \Delta t \leq \Delta x$~\cite{goatin2023macroscopic}, we establish the time interval and length of the road cell as $\Delta t = 1$ second and $\Delta x = 50$ meters, respectively, with the maximum vehicle speed set to $V = 15$ m/s. Consequently, the total number of road cells is 16, resulting in a state vector comprising 34 dimensions.

The entire road section is divided into 8 segments, each containing two cells and spanning 100 meters in length. For each segment, two parameters are defined: the maximum density $R$ and the maximum speed $V$. Initially, the maximum speed for all segments is uniformly set to $15$ m/s, consistent with the state space. The initial maximum density for segments with three lanes is set to 20 per cell, while two-thirds of this value is assigned to segments with two lanes. The covariance matrix $P$ is initialized as an identity matrix with corresponding dimensions.

For Dyna-Q learning, we utilize an Adam optimizer with a learning rate of 0.0001 to update the Q network. The size of the replay buffer is set to 300,000, where new experiences replace old ones when the buffer is full. During each training iteration, batches of experiences with a size of 64 are sampled from the replay buffer. Actions are chosen according to an $\epsilon$-greedy policy. Initially, within the first 20,000 training steps, $\epsilon$ gradually increases to its maximum value of 0.95. The discount factor is set to 0.99.

\subsection{Evaluation of the PDE-ODE model}

We assess the the PDE-ODE model's accuracy in two key aspects: density prediction and speed prediction. We enable the trained platoon to interact with the SUMO environment and compare the model's predictions with the true states, recording the error at each time step. This evaluation allows us to gauge the model's performance and identify areas for improvement.

\begin{figure}[htbp]
    \centering
    \includegraphics[width=0.5\textwidth]{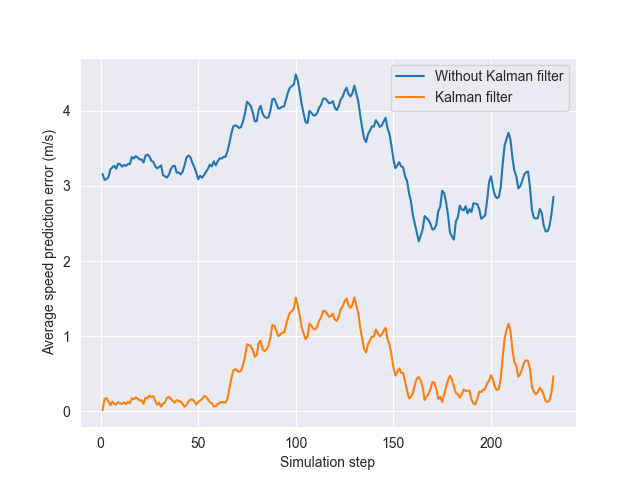}
    \caption{Average speed prediction errors for all cells using Kalman filter-trained parameters and untrained parameters.}
    \label{state_error}
\end{figure}

Utilizing our approach, the agent spends over 200 time steps interacting with the SUMO environment. Table~\ref{Diff_state_avg_table} presents the average prediction errors for all cells throughout the simulation process. Notably, the density prediction proves remarkably accurate, with the average absolute error between predicted and true density remaining under $0.12$ veh/cell. Interestingly, the density error slightly decreases towards the end of the episode compared to the beginning. Furthermore, we examine the speed prediction error for both untrained parameters and those updated with the Kalman filter, depicted in Fig.~\ref{state_error}. Notably, the error peaks around $100$ and $200$ time steps for both parameter sets, albeit significantly higher for untrained parameters. Throughout the episode, the absolute speed error remains below $1.5$ m/s for parameters updated with the Kalman filter, whereas untrained parameters consistently yield errors exceeding $2$ m/s, with some instances even surpassing $4$ m/s. Clearly, post-training, the speed prediction error undergoes substantial reduction, signifying a marked improvement in model accuracy.

\begin{table}[htbp]
\centering
    \begin{tabular}{ c | c | c }
        \hline
         & Absolute error & Relative error \\
        \hline
        Density & 0.12 veh/cell & 0.6\% \\
        \hline
        Speed & 1.5 m/s & 10\% \\
        \hline
    \end{tabular}
\caption{Average absolute and relative prediction errors.}
\label{Diff_state_avg_table}
\end{table}

The speed prediction, although not as accurate as density prediction, remains a valuable tool in our approach. This discrepancy arises from the limitations of using a first-order linear model to capture the relationship between density and speed, rendering it unable to fully capture complex road conditions such as traffic jams. However, we demonstrate that despite this limitation, an approximately accurate model can still facilitate the training of the agent towards learning a more effective control policy.

\subsection{Performance comparison}

\begin{figure}[htbp]
\centering
\includegraphics[width=0.5\textwidth]{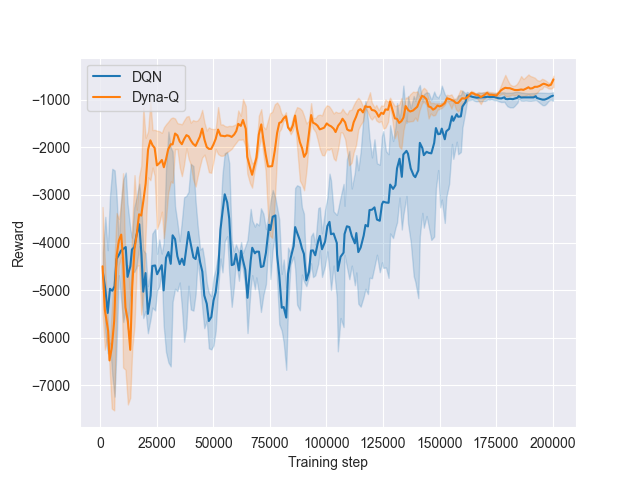}
\caption{A comparison between the reward curves of the DQN and Dyna-Q methods, where Dyna-Q incorporates the macroscopic model.}
\label{DynaQ_cmp}
\end{figure}

By integrating the PDE-ODE model updated through the Kalman filter with DQN, we develop a model-based algorithm akin to Dyna-Q. To enable a comprehensive comparison, we utilize the original DQN as a baseline method. Both approaches are employed to train agents for $200$ thousand steps across 3 random seeds. The training curves of both methods are depicted in Fig.~\ref{DynaQ_cmp}.
Rewards lower than $-3000$ indicate that the agent is penalized for timeouts. From this figure, we observe that the agent trained with Dyna-Q starts to mitigate these penalties at around 20,000 training steps. Subsequently, the Dyna-Q agent gradually converges to a policy characterized by higher rewards and lower fuel consumption. Conversely, the DQN agent fails to converge to a stable policy within 100,000 training steps. This illustrates that our algorithm, incorporating the density-speed model, enhances training efficiency and facilitates faster convergence towards an optimal control policy.

\begin{figure}[htbp]
    \centering
    \includegraphics[width=0.5\textwidth]{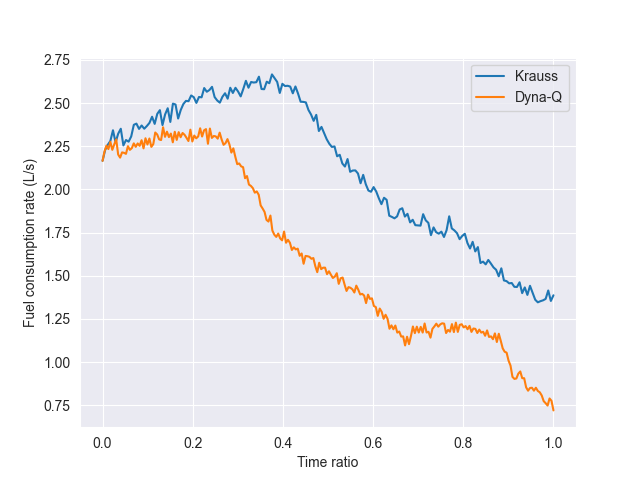}%
    \caption{Fuel consumption rates throughout the simulation under the Krauss model and Dyna-Q.}
    \label{costCmp}
\end{figure}

The Krauss model serves as a typical car-following model in SUMO settings \cite{pourabdollah2017calibration}. In our setup, we employ the Krauss model to control the vehicle platoon and consider it as a benchmark. The total fuel consumption rates of the two control methods are depicted in Fig~\ref{costCmp}. Throughout the entire driving process, the fuel consumption rate of the Dyna-Q policy control consistently remains lower than that of the Krauss model control. Specifically, the rewards for the Dyna-Q agent and Krauss model are respectively $-496.07$ and $-1052.04$. Excluding the reward for entering the next cell and acceleration penalty, the total fuel costs through conversion for Dyna-Q policy control and Krauss model control are respectively $395.60$ L and $440.12$ L, resulting in a reduction in fuel consumption by approximately $10.11\%$.

\begin{figure*}[htbp]
    \centering
    \subfigure[Control with the Krauss model]{%
    \label{DensChange: up}
        \includegraphics[width=0.45\textwidth]{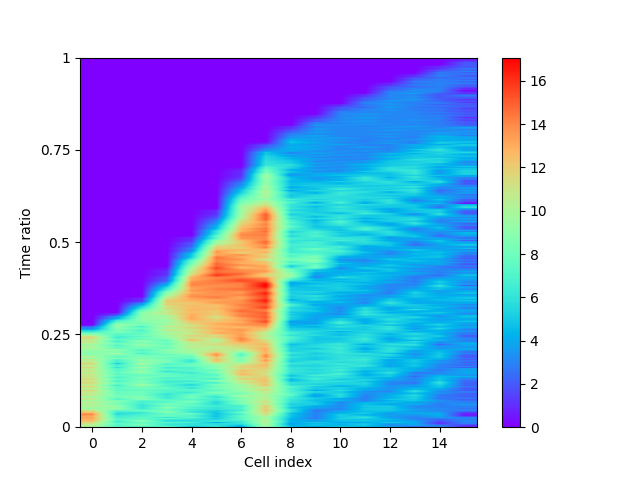}%
    }
    \hfill
    \subfigure[Control with our Dyna-Q framework]{%
    \label{DensChange: down}
        \includegraphics[width=0.45\textwidth]{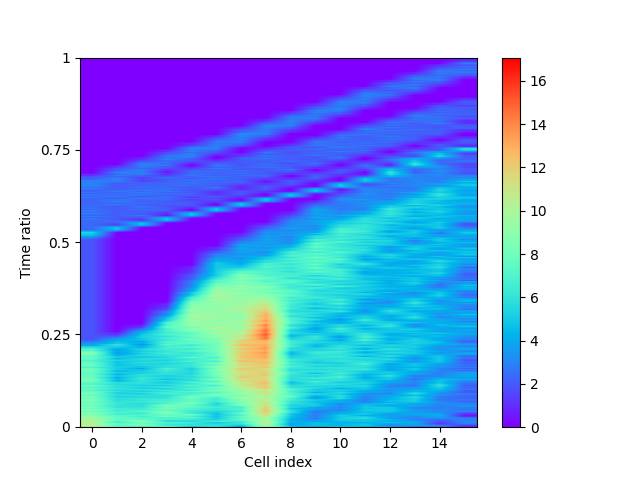}%
    }
    \caption{Traffic density profiles under two control methods.}
    \label{DensChange}
\end{figure*}

Traffic density profiles under both methods are illustrated in Fig~\ref{DensChange}. In Fig.\ref{DensChange: up}, the Krauss model control leads to a significant congestion scenario. Notably, cells numbered from 4 to 7, corresponding to the road section between $200$ m and $350$ m, exhibit densities surpassing 10 vehicles per cell within the initial $65\%$ of the total time, with a peak density reaching about 17 vehicles per cell. Conversely, Fig.\ref{DensChange: down} showcases a notable reduction in congestion length and duration under the Dyna-Q policy control. Here, congestion only persists between cell 6 and cell 7 within the initial $30\%$ of the total time, with a peak density of approximately 14 vehicles per cell—lower than observed under the Krauss model. These results underscore the capacity of the Dyna-Q policy not only to curtail total fuel consumption but also to alleviate congestion upstream of the bottleneck.

\section{Concluding remarks} \label{sec:conclusion}

This paper introduces a novel approach to reduce fuel consumption of vehicles at highway bottlenecks in mixed autonomy traffic through the control of vehicle platoons. We propose a Dyna-style learning framework integrating a coupled PDE-ODE model. Additionally, we enhance the PDE-ODE model's predictive capabilities with a Kalman filter to estimate density and average speed for each road cell. Using the SUMO simulation, we evaluate the efficacy of our approach. Furthermore, we assess the performance of the Dyna-style planning and learning against traditional car-following methods. Our results demonstrate a reduction in total fuel consumption by approximately $10\%$. Moreover, our method exhibits faster convergence to an optimal policy compared to model-free learning, owing to the incorporation of the macroscopic model.

This work can be extended in several directions. Firstly, considering the platoon as a rigid body with a fixed length suggests the potential for adopting a more realistic platoon model in the simulation. Additionally, the first-order traffic model employed here may be too simplistic for complex scenarios, indicating the need for more precise traffic models. Finally, future research could explore the utilization of advanced deep reinforcement learning algorithms beyond simple Dyna-Q.

\bibliographystyle{unsrt}  
\bibliography{DYNA_STYLE}
\end{document}